\documentclass[runningheads]{llncs}
\usepackage[T1]{fontenc}
\usepackage{graphicx}
\usepackage{booktabs}
\usepackage[misc]{ifsym}

\usepackage{mwe}

\usepackage{times}
\usepackage{soul}
\usepackage{url}
\usepackage[hidelinks]{hyperref}
\usepackage[utf8]{inputenc}
\usepackage[small]{caption}
\usepackage{graphicx}
\usepackage{amsmath}
\usepackage{breqn}
\usepackage{booktabs}
\usepackage{algorithm}
\usepackage{ifpdf}
\usepackage{subcaption}

\usepackage{algorithmic}
\usepackage{comment}
\usepackage{multirow}
\usepackage{latexsym}
\usepackage{cite}
\usepackage{colortbl}

\usepackage{amssymb,bm}

\begin{document}

\title{Graph Neural Network leveraging Higher-order Class Label Connectivity for Heterophilous Graphs}

\titlerunning{GNN leveraging Higher-order Class Label Connectivity}

\toctitle{Graph Neural Network leveraging Higher-order Class Label Connectivity for Heterophilous Graphs}
\tocauthor{Takuto Takahashi, Itsuki Nakayama, Takahiro Mitani, Ryosuke Kikuchi, Yuya Sasaki, Makoto Onizuka}


\author{Takuto Takahashi \Letter \and
Itsuki Nakayama \and
Takahiro Mitani \and
Ryosuke Kikuchi \and
Yuya Sasaki \and
Makoto Onizuka
}


\authorrunning{T. Takahashi et al.}
\institute{The University of Osaka, 1-5 Yamadaoka, Suita, Osaka, Japan \email{\{takahashi.takuto, nakayama.itsuki, mitani.takashiro, kikuchi.ryosuke, sasaki, onizuka\}@ist.osaka-u.ac.jp}}

\maketitle              

\begin{abstract}
Node classification in graph neural networks (GNNs) has been widely applied in various fields of graph analysis. 
GNNs achieve high-accuracy node classification in homophilous graphs, where nodes with the same class label tend to be connected. 
However, their performance remains limited in heterophilous graphs, where nodes with different class labels are more likely to be connected.
In particular, current GNNs derived from graph convolutional networks cannot capture higher-order class label connectivity, which is frequently observed in real-world heterophilous graphs. 
To address this issue, we propose a novel classifier, Label Context Classifier (LCC), designed to capture higher-order class label connectivity in directed graphs. 
LCC estimates the class label of a target node by leveraging {\it label context} embeddings that are generated through four distinct types of walks.
In addition, our approach allows the integration of LCC and any GNN by adaptively learning their importance.
Experimental results demonstrate that GNNs integrated with LCC outperform SOTA methods and the label context embeddings improve the node classification performance in heterophilous directed graphs.

\end{abstract}

\section{Introduction}
\label{sec:intro}
Node classification in graphs is one of the important tasks in graph analysis, aiming to predict the class labels of nodes. 
This task has a wide range of applications, including the analysis of social networks and biological networks, such as genes and proteins~\cite{ReviewofMethodsandApplications, Molecular, Physics, science}.
A representative approach for node classification is Graph Neural Networks (GNNs)~\cite{GCN,GAT,JKNet,APPNP,GraphSAGE,GraphSAINT,H2GCN, LINKX,CAGNNs,LG-GNN,maekawa2022beyond}.
Traditional GNNs such as Graph Convolutional Network (GCN)~\cite{GCN} are designed for homophilous graphs, where nodes with the same class labels/features are more likely to be connected. 
However, their effectiveness is limited for heterophilous graphs, where nodes with different class labels/features tend to be connected~\cite{GNNheterophily, HeterophilyGraph}.
To improve the performance in heterophilous graphs, GNNs that capture the characteristics of heterophilous graphs have been actively proposed~\cite{H2GCN, GNNheterophily, LINKX, GloGNN, RevisitingHeterophily, CAGNNs, LG-GNN}.
Nevertheless, there are still cases where the accuracy of these GNNs does not surpass that of multilayer perceptrons (MLPs), which rely only on node features without using edges. This result indicates that these GNNs do not fully leverage the structural information of graphs~\cite{RevisitingHeterophily}.

\noindent
{\bf Motivation.}
An interesting observation we found on the weakness of current GNNs derived from GCN is that they fail to capture the {\it higher-order class label connectivity} in graphs, that is, how class labels are connected through multiple hops using directed edges.
This weakness is caused by the fact that GCN transforms the node classification problem into a simple classification problem using graph convolution operation: after applying the graph convolution operations, GCN trains the model from embedding space to class label space, so it ignores the class label connectivity among nodes, particularly appeared in the training sets.
This limitation also applies to more advanced methods such as \cite{GNNheterophily, DBLP:journals/tmlr/ZhongIP22}. While they utilize a class compatibility matrix corresponding to 1st-order class label connectivity, they do not capture higher-order class connectivity.



\begin{figure}[t]
    \centering
    \begin{minipage}{0.27\textwidth}
        \centering
        \includegraphics[width=\textwidth]{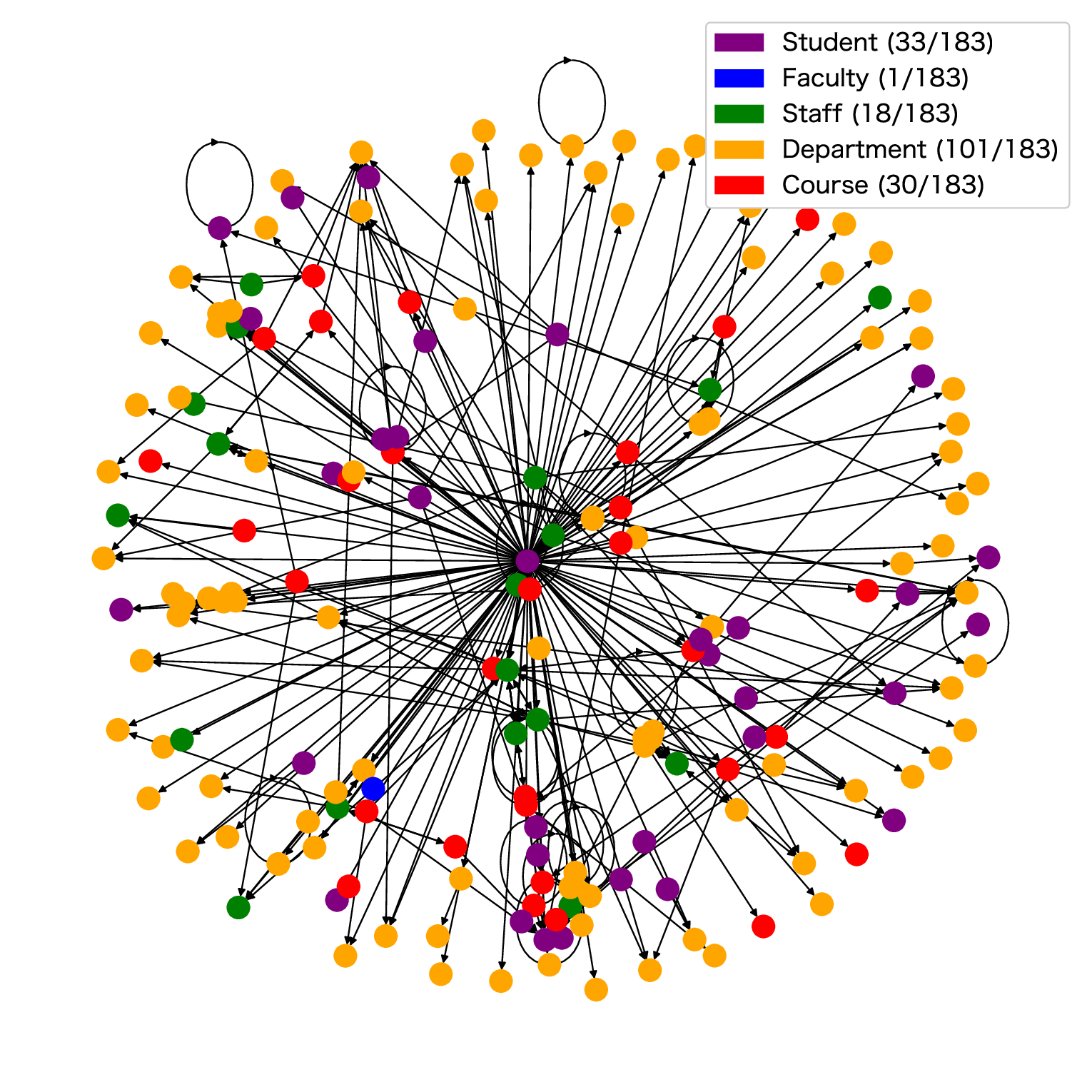}
        \vspace{-5mm}
        \subcaption{Graph visualization}
    \end{minipage}
    \begin{minipage}{0.345\textwidth}
        \centering
        \includegraphics[width=\textwidth]{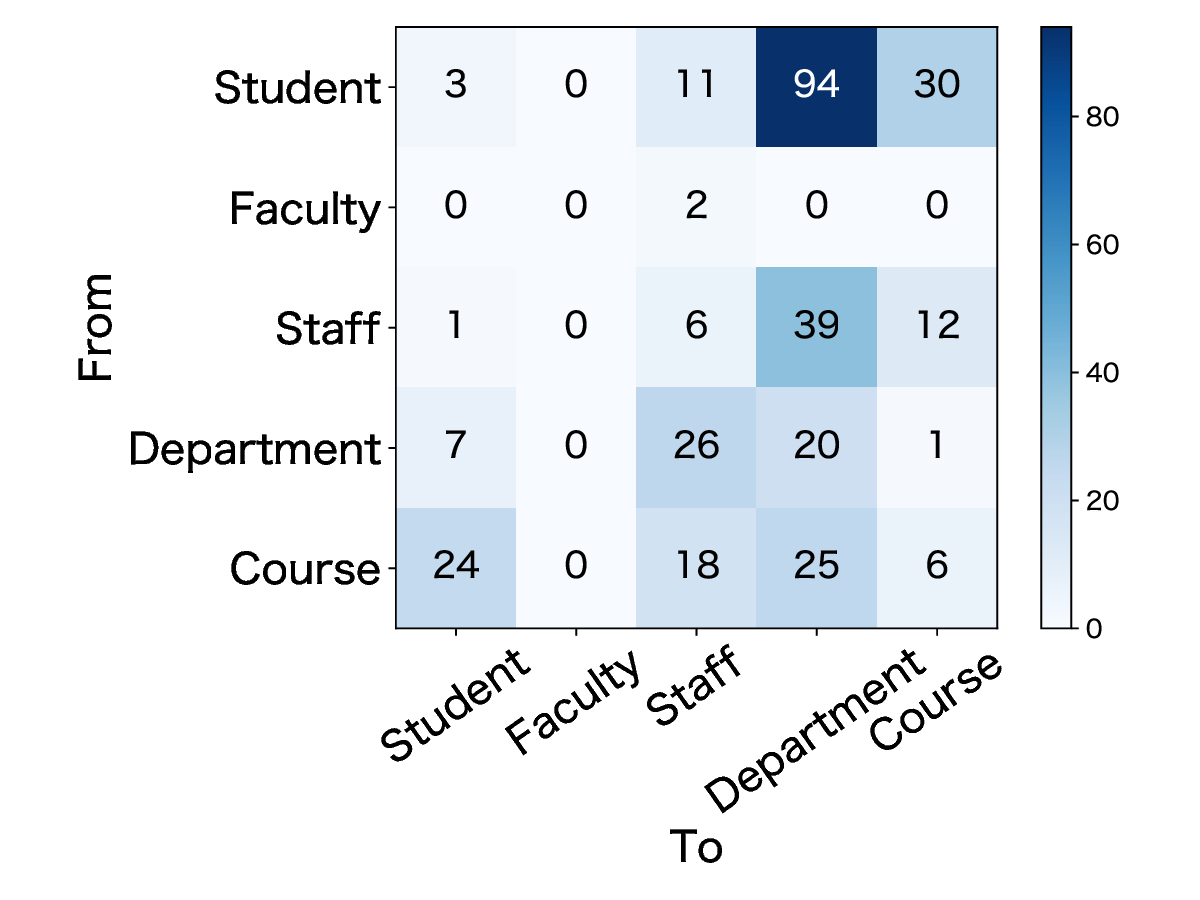}
        \subcaption{1st-order connectivity}
    \end{minipage}
    \begin{minipage}{0.36\textwidth}
        \centering
        \includegraphics[width=\textwidth]{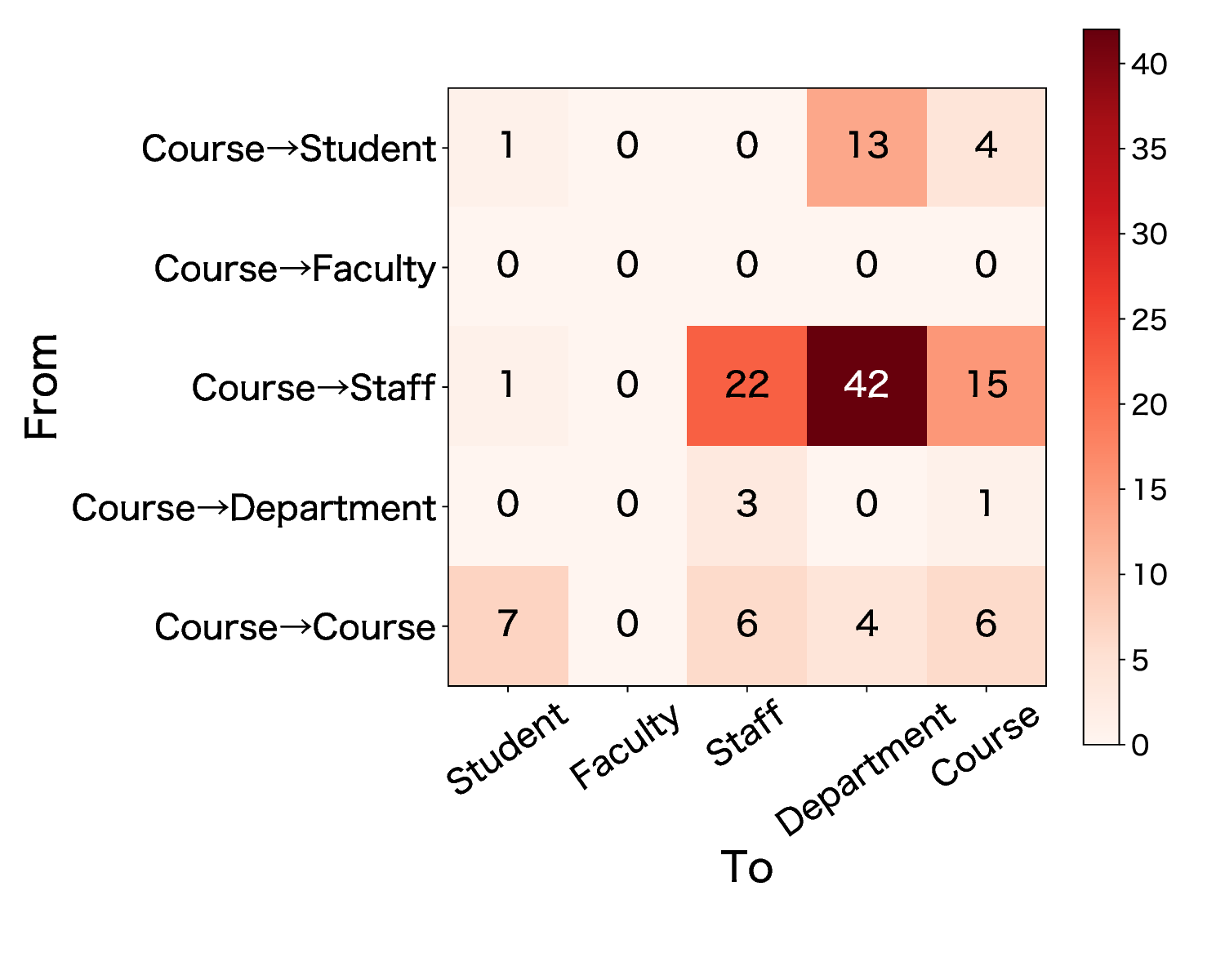}
        \vspace{-7mm}
        \subcaption{2nd-order connectivity: we only show five rows of Course$\rightarrow$* paths out of 25 rows}
    \end{minipage}

    \caption{The class label connectivity in the Texas dataset. (a) visualization with five class labels. (b) 1st-order connectivity (\#edges) from class label in y-axis to class label in x-axis. (c) 2nd-order connectivity from class label path (e.g., Course$\rightarrow$Student) in y-axis to class label in x-axis.}
    \label{fig:texas_edgecount}
\end{figure}

Strong higher-order class label connectivity often appears in real-world graphs.
Figure~\ref{fig:texas_edgecount} illustrates such examples in the heterophilous directed graph of the Texas dataset.
Figure~\ref{fig:texas_edgecount} (a) visualizes the graph with five class labels (i.e., Student, Faculty, Staff, Department, and Course).
Figure~\ref{fig:texas_edgecount} (b) shows the 1st-order connectivity, depicting the number of directed edges from the class label on the y-axis to the one on the x-axis. We observe that there is strong connectivity from Student nodes to Department nodes, whereas there is no connectivity to Faculty nodes.
Similarly, Figure~\ref{fig:texas_edgecount} (c) also reveals strong/weak 2nd-order class label connectivity. 
For example,  the connectivity from Course$\rightarrow$Staff class label path to Department nodes is strong, whereas no connectivity to Faculty nodes.

\noindent
{\bf Contribution.}
In order to capture such higher-order class label connectivity, 
we propose a novel classifier, Label Context Classifier (LCC), 
which trains the model using {\it label walks}, which are sequences of class labels on walks.
First, we extract four fundamental types of label walks from a target node: forward walk, backward walk, sibling walk, and guardian walk. 
They are mutually exclusive, and each walk type captures a different aspect of class label connectivity.
We generate {\it label context embeddings} that capture the label context by training a model using the label walks, an idea inspired by word2vec~\cite{word2vec}.
Then, we train LCC using the concatenation of node features and the label context embeddings obtained from different types of label walks.
Since LCC complements the capability of GNNs, we integrate LCC and any GNN by adaptively learning their importance using validation loss without additional model training.

The contributions of this paper are as follows:
\begin{itemize}
    \item We propose the Label Context Classifier (LCC), which estimates the class label of a target node by capturing the higher-order class label connectivity across directed graphs.
    LCC estimates the class label using the label context embeddings generated from four types of label walks.
    \item 
    We can integrate LCC and any GNN by adaptively learning their importance weights without additional model training.
    \item Experimental results confirm that our proposal outperforms SOTA methods and the label context embeddings actually enhance node classification performance in heterophilous directed graphs.
\end{itemize}
The structure of this paper is as follows.
Section \ref{sec:related} describes related work, and then Section \ref{sec:preliminary} explains preliminary knowledge.
Sections \ref{sec:lcc} and \ref{sec:integration} present the details of the Label Context Classifier and how to integrate it and any GNN, respectively.
Section \ref{sec:experiments} conducts experiments to demonstrate the effectiveness of the proposed method for heterophilous directed graphs.
Finally, Section \ref{sec:conclusion} concludes this paper.

\section{Related Work}
\label{sec:related}

\subsubsection*{Traditional GNNs}
Since Graph Convolutional Network (GCN)~\cite{GCN} has emerged, numerous methods have been proposed for node classification using graph convolution. The graph convolution ensures 1st-order node proximity, which makes the adjacent node representations similar. Therefore, a family of GCN is suitable for homophily graphs but not for heterophilous graphs. 

Graph Attention Networks (GAT)~\cite{GAT} improve accuracy by employing a self-attention mechanism that learns relative weights between connected node pairs.
JK-Net~\cite{JKNet} is a method that aggregates outputs from multiple layers of a GNN to integrate information at different layer levels.
APPNP~\cite{APPNP} first transforms the initial node features using an MLP and then applies a personalized PageRank-based iterative propagation mechanism to distribute information among nodes.
In addition, sampling-based GNNs~\cite{GraphSAGE, GraphSAINT} are scalable methods that compute node representations using subgraphs extracted from the input graph.
GraphSAGE~\cite{GraphSAGE} samples a fixed number of neighbors uniformly for each node.
GraphSAINT~\cite{GraphSAINT} samples subgraphs and learns graph representations by combining information from multiple subgraphs.

\subsubsection*{GNNs for Heterophilous Graphs}
Real-world graphs sometimes exhibit heterophily, where nodes of different attributes/classes are more likely to be connected. 
There have been several GNNs designed for heterophilous graphs~\cite{H2GCN,LINKX,GloGNN,CAGNNs,LG-GNN,maekawa2023simple}.
However, these methods also suffer from the same limitation as the GNNs derived from GCN in the sense that they also transform the node classification problem into a simple classification problem by training the model from embedding space to class label space, so they ignore the class label connectivity.

H2GCN~\cite{H2GCN}, a representative GNN designed for heterophilous graphs, separates the processing of a node’s own features from those of its neighbors, utilizes high-order neighborhood information, and employs an appropriate aggregation function to capture complex relationships between nodes. 
LINKX~\cite{LINKX} independently processes node features and adjacency matrix information, learning both in parallel to effectively leverage multiple sources of information.
GloGNN~\cite{GloGNN} combines local node features with global features that capture relationships between distant but structurally similar nodes.
Recent studies have further advanced this area. CAGNNs~\cite{CAGNNs} introduce a novel metric based on the distinguishability of neighboring nodes, decomposing node features into representation and aggregation components. A mixer module is then used to adaptively evaluate neighboring information for each node.
Adaptive Channel Mixing (ACM)~\cite{RevisitingHeterophily} is a GNN framework designed to address heterophily by adaptively combining aggregation, diversification, and identity channels at the node level. ACM allows nodes to learn different weights for each channel, effectively capturing local heterophily without requiring high-order filters or increased computational resources.
LG-GNN~\cite{LG-GNN} achieves high-accuracy node classification for heterophilous graphs by adaptively integrating global structural similarity and local feature similarity between nodes. This design effectively considers node relationships during the information aggregation and propagation process.

\subsubsection*{Other related work}
There are methods~\cite{GNNheterophily,DBLP:journals/tmlr/ZhongIP22} that utilize the class compatibility matrix, which corresponds to 1st-order class label connectivity. 
As a specific example, CPGNN~\cite{GNNheterophily} trains the class compatibility matrix using training sets and a prior belief estimator. However, it has a weakness in that it does not capture higher-order class connectivity.

A meta-path is a predefined sequence of node and edge types in order to capture higher-order connectivity and semantic relationships between different entities in heterogeneous information networks (HIN).
For example, MetaPath2Vec~\cite{metapath2vec} generates constrained random walks for learning embeddings, while HAN~\cite{HAN} employs attention mechanisms to aggregate multiple meta-paths. Recent studies~\cite{Meta-Path-Discovery} focus on automatic meta-path discovery to enhance model generalization.
These methods assume that all nodes and edges are typed, and meta-paths are defined based on those types. In contrast, our approach is applicable to graphs without predefined types, and we define four types of direction-aware fundamental walks.

node2vec~\cite{node2vec} is a method to generate node embedding in order to capture homophily and structural equivalence. 
Our proposal and node2vec share some common characteristics, such as generating walks and using word2vec. However, there are two major differences.
First, node2vec does not use class labels as input and is not designed for directed graphs. 
Second, the type of walk is limited to only a single type, and the parameters that determine the balance between breadth-first and depth-first search must be manually set by the user. 
In contrast, in order to capture class connectivity, we define four types of direction-aware class label walks, and their importance is learned automatically without manual intervention. 
\section{Preliminary}
\label{sec:preliminary}

\subsubsection*{Graph}
We consider a directed graph $G = (V, E)$, which consists of a node set $V$ with $n$ nodes and a directed edge set $E$ with $m$ edges. 
The adjacency matrix $\mathbf{A} \in \{0, 1\}^{n \times n}$ is defined such that $a_{ij} = 1$ if $(v_i, v_j) \in E$, and $a_{ij} = 0$ otherwise. 
Additionally, we define the feature matrix $\bm{X} \in \mathbb{R}^{n \times d}$, where each node is assigned a $d$-dimensional feature vector. 
Each node $v$ has a unique class label $y_v \in \{1, \dots, C\}$ (number of classes: $C$), and the class label vector is denoted as $\mathbf{y}$. 
For clarity in explanations, we refer to the starting node of a directed edge as the parent node and the ending node as the child node. Additionally, we use the terms, sibling nodes and guardian nodes, which are naturally defined based on the parent-child relationship between nodes.

We define heterophilous graphs after defining edge homophily~\cite{H2GCN}.
The edge homophily~\cite{H2GCN} is calculated as follows.
\begin{equation}
    \textstyle H(G) = \frac{\sum_{0 \leq i, j < n} a_{ij} \delta(y_{v_i}, y_{v_j})}{m},
\end{equation}
where $\delta(y_{v_i}, y_{v_j})$ returns one if $y_{v_i} = y_{v_j}$ otherwise zero.
We define graphs with low $H(G)$  as heterophilous graphs.
A lower edge homophily indicates a stronger tendency toward heterophily.

\subsubsection*{Problem Definition (Node classification)}
We split a node set $V$ into a training set $V_{train}$, validation set $V_{val}$, and test set $V_{test}$.
Given adjacency matrix $\mathbf{A}$, feature matrix $\bm{X}$, and node class labels in $V_{train}$ and $V_{val}$, we predict the labels of the nodes in $V_{test}$. 

\section{Label Context Classifier}
\label{sec:lcc}
GNNs fail to capture the higher-order class label connectivity, which often appears in real-world graphs, as we described in Section \ref{sec:intro}.
This weakness is caused by the fact that GNNs transform the node classification problem into a simple classification problem:  GNNs train the model from embedding space to class label space, so they ignore the class label connectivity among nodes, particularly appeared in the training sets.

To this end, 
we propose a novel classifier, Label Context Classifier (LCC), which trains the model using various types of label walks in order to capture higher-order class label connectivity.
\begin{figure*}[t] 
    \centering 
        \centering
        \includegraphics[width=\textwidth]{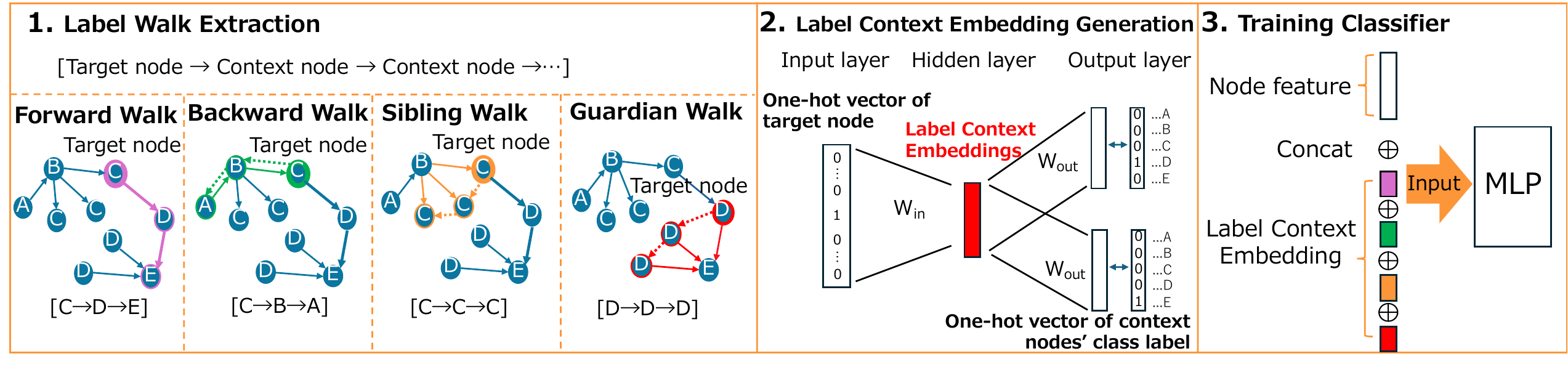}
    \caption{
    The framework of Label Context Classifier (LCC). LCC consists of three steps: 1) extraction of four types of label walks, 2) generation of label context embeddings, and 3) training a classifier using the label context embeddings.
    }
    \label{fig:lcc} 
\end{figure*}
Our method consists of the following steps as illustrated in Figure~\ref{fig:lcc}: 
(1) extract four types of fundamental label walks: a simple walk (forward walk, backward walk) and a mixture of forward and backward walks (sibling walk, guardian walk). They are mutually exclusive and each walk type captures a different aspect of class label connectivity,
(2) generate embeddings that capture the label context by training a model using the label walks, and
(3) train LCC using the concatenation of node attributes and the label context embeddings obtained from different types of label walks. This enables the model to appropriately select suitable embeddings for node classification.

\subsection{Label Walk Extraction}
A label walk is defined as a walk obtained from a graph where the nodes in the walk are replaced with their labels in the training set. 
We refer to the first label in the walk as the target label and the remaining labels as context labels.
If a node in the walk is not in the training dataset, we use the null label in the label walk, and it is ignored in the label context embedding step.

Table~\ref{tbl:walks} presents a summary of the characteristics of the four label walk types, indicating they are mutually exclusive and each walk type captures a different aspect of higher-order class label connectivity.
Step $1$ in Figure~\ref{fig:lcc} shows examples of label walk types: forward walk $(C, D , E)$, backward walk $(C , B , A)$, sibling walk $(C , C , C)$, and guardian walk $(D , D , D)$.
The details are described in the following sections.

\begin{table}[t]
    \centering
    \caption{Summary of label walk types and their characteristics.}
    \begin{tabular}{l||l|l|c}
        \hline
        \multicolumn{1}{c||}{\textbf{Walk type}} & \multicolumn{1}{c|}{\textbf{Description}} & \multicolumn{1}{c|}{\textbf{Traverse}} & \multicolumn{1}{c}{\textbf{Class label connectivity}} \\
         &  & \multicolumn{1}{c|}{\textbf{edge direction}} & \multicolumn{1}{c}{\textbf{order (\#hops)}} \\
        \hline
        Forward   & Depth-first search-based walk & forward only & walk length \\
        Backward  & Inverse of forward walk & backward only & walk length \\
        Sibling   & Walk on sibling nodes & backward + forward & 2 \\
        Guardian  & Inverse of sibling walk & forward + backward & 2 \\
        \hline
    \end{tabular}
    \label{tbl:walks}
\end{table}

\subsubsection{Forward Walk}
The forward walk aims to capture the higher-order class label connectivity directed from a target label down to neighboring context labels. 
We extract forward walks using the depth-first search that follows the forward direction of directed edges. For the given target node $v_0$ and the length $w$ of label walk, a forward walk (FW) is formulated as follows:
\begin{equation}
\begin{aligned}
FW(v_0) &= (y_{v_0}, y_{v_1}, \dots, y_{v_{w}}), 
&\quad \text{s.t.} (v_i, v_{i+1}) \in E, i\in\{0,1, \dots, w-1\}.
\end{aligned}
\end{equation}
The order of class label connectivity is walk length. For example, it is 1st-order if the walk length is 1 (i.e., 1 hop).

Figure~\ref{fig:texas_edgecount} shows an example: the Student nodes are more likely to connect to the Department nodes (1st order connectivity), and those Department nodes are more likely to connect to the Staff nodes (2nd order connectivity).

\subsubsection{Backward Walk}
The backward walk is the inverse notion of the forward walk.
The reason why we introduce the backward walk in addition to the forward walk is that the edge direction is user-defined, so both directions are useful.
Also, the backward walk is useful for the sink nodes (i.e., nodes without outgoing edges), because they cannot utilize forward walks. 
We extract backward walks using the depth-first search that follows the reverse direction of directed edges.
Similar to the forward walk, a backward walk (BW) is defined as follows:
\begin{equation}
\begin{aligned}
BW(v_0) &= (y_{v_0}, y_{v_1}, \dots, y_{v_{w}}),
&\quad \text{s.t.} (v_{i+1}, v_i) \in E, i\in\{0,1, \dots, w-1\}.
\end{aligned}
\end{equation}

Figure~\ref{fig:texas_edgecount} shows an example: the Department nodes are more likely to be connected from the Student nodes (1st order connectivity), and those Student nodes are more likely to be connected from the Course nodes (2nd order connectivity).

\subsubsection{Sibling Walk}
In addition to the simple forward/backward walks, we introduce the sibling walk, our new idea which is a mixture of forward and backward walks. 
The motivation for introducing sibling walks is that sibling nodes (the child nodes connected to the same parent node) often share the same class label for certain target nodes. 
We extract sibling walks by 1) traversing backward to a parent node of the target node $v_0$, and then 2) repeatedly traversing forward to its child nodes until reaching the desired label walk length $w$.
A sibling walk (SW)\footnote{Since a sibling walk traverses multiple sibling nodes without edges, it does not strictly follow the ``walk'' definition in the graph theory.} is formulated as follows:
\begin{equation}
\begin{aligned}
SW(v_0) &= (y_{v_0}, y_{v_1}, \dots, y_{v_{w}}) \\
&\quad \text{s.t.} \quad \exists p \in \text{parents}(v_0), \quad 
v_1, \dots, v_{w} \in \text{children}(p) \setminus \{v_0\}.
\end{aligned}
\end{equation}
%
where $\text{parents}(v)$ and $\text{children}(v)$ are parent nodes and child nodes of $v$, respectively. 
The order of class label connectivity is 2 regardless of the walk length, because the target node and the sibling nodes are 2-hops apart.

Figure~\ref{fig:texas_edgecount} shows an example: the sibling walk captures the connectivity between the nodes labeled as Student whose parent nodes are labeled as Course.

\subsubsection{Guardian Walk}
The guardian walk is the inverse notion of the sibling walk.
Similarly to the sibling walk, the motivation for introducing guardian walks is that guardian nodes often share the same class label for certain target nodes. 
Compared to the general walks based on depth-first or breadth-first search, the sibling walk and guardian walk extract only siblings and guardians, which mitigates the noisy effect on the downstream classifier. Indeed, our experiments in Section \ref{sec:experiments} verify that the sibling walk and guardian walk significantly improve the accuracy. 
We extract guardian walks by 1) traversing forward to a child node of the target node $v_0$, and then 2) repeatedly traversing backward to its parent nodes until reaching the desired label walk length $w$.
A guardian walk (GW) is defined as follows:
\begin{equation}
\begin{aligned}
GW(v_0) &= (y_{v_0}, y_{v_1}, \dots, y_{v_{w}}) \\
&\quad \text{s.t.} \quad \exists c \in \text{children}(v_0), \quad 
v_1, \dots, v_{w} \in \text{parents}(c) \setminus \{v_0\}.
\end{aligned}
\end{equation}
%
Figure~\ref{fig:texas_edgecount} shows an example: the guardian walk captures the connectivity between the nodes labeled as Student whose child nodes are labeled as Department.

\begin{algorithm}[t]
\caption{Label context embedding matrix generation}
\label{algo:Labelembeddings}
\begin{algorithmic}[1] 
\REQUIRE adjacency matrix $\mathbf{A}$,  class label $\mathbf{y}$, embedding dimension $d'$, label walk length $w$, \\number of label walk $k$, epoch $T$
\ENSURE label context embedding matrix $\mathbf{Z} \in \mathbb{R}^{n \times d'}$
\STATE {\bf \#\#\# Initialize \#\#\#}
\STATE Initialize the label context embedding $\mathbf{z}_v \in \mathbb{R}^{d'}$ of node $v \in V$ to a random value
\STATE Create  an one-hot vector $\mathbf{\ell}_v$ of class label for node $v$ from $\mathbf{y}$

\STATE {\bf \#\#\# Extract label walks \#\#\#}   
\FOR{ $v \in V$}
    \IF{label walk is forward / backward walk}
        \FOR{$i = 1, \dots, k$}
            \STATE Extract label walks (forward, backward walks) for $v$
        \ENDFOR
    \ELSIF{label walk is sibling / guardian walk}
        \STATE Extract a label walk (sibling, guardian walk) for $v$
    \ENDIF
\ENDFOR

\STATE {\bf \#\#\# Train the model for label context embedding \#\#\#}
\FOR{$t=1, \dots,T $}
    \FOR{each label walk}
        \STATE Calculate output embedding $\hat{\mathbf{\ell}}_v= \mathbf{z}_v^\top \mathbf{W_{out}}$
        \STATE Calculate cross entropy as loss $\mathcal{L}= \sum_{\mathbf{\ell}_u \in contexts(v)}\mathcal{L}_{CE}(\hat{\mathbf{\ell}}_v, \mathbf{\ell}_u)$
        \STATE Update $\mathbf{z}_v$, $\hat{\mathbf{y}}_v$ and $\mathbf{W_{out}}$ to minimize $\mathcal{L}$
    \ENDFOR
\ENDFOR

\STATE {\bf \#\#\# Output label context embedding matrix \#\#\#}
\RETURN $\mathbf{Z} = [\mathbf{z}_v]_{v \in V}$ 
\end{algorithmic}
\end{algorithm}

\subsection{Label Context Embeddings}
We generate embeddings that capture the label context using the label walks.
Our purpose is to verify the effectiveness of leveraging label walks for capturing higher-order class label connectivity, so we use a simple two-layer MLP to train the model to predict the context labels using the target node as input. The idea is inspired by the Skip-gram model of word2vec~\cite{word2vec}. Extending our framework to use more advanced techniques, such as transformers, is part of future work. 
Specifically, the target node is represented as a one-hot node vector in the input layer, and the context labels are represented as a one-hot label vector in the output layer. 
The model is trained to minimize the loss between the estimated output and the ground-truth context labels, 
ensuring that the internal layer represents the embedding for a given input node.

Step 2 in Figure~\ref{fig:lcc} shows an overview of generating label context embeddings. 
Let $\mathbf{Z} \in \mathbb{R}^{n \times d'}$ be the label context embedding matrix obtained in the hidden layer for each node where $d'$ represents the dimension of the embedding vector.
The output layer embedding $\hat{\mathbf{\ell}}_v$ is obtained by multiplying the label context embedding $\mathbf{z}_v$ of the target node $v$ by a weight matrix $\mathbf{W_{out}} \in \mathbb{R}^{d' \times C}$.
\begin{equation}
    \hat{\mathbf{\ell}}_v = \mathbf{z}_v^\top \mathbf{W_{out}}
\end{equation}
Then, the cross-entropy loss $\mathcal{L}_{CE}$ is computed with respect to the one-hot vector $\mathbf{\ell}_u$ of the ground-truth context label.
The total loss $\mathcal{L}$ over all pairs of $\hat{\mathbf{\ell}}_v$ of target node $v$ and its context label $\mathbf{\ell}_u$ is formulated as follows:
\begin{equation}
    \mathcal{L} = \sum_{\ell_u \in contexts(v)}\mathcal{L}_{CE}(\hat{\mathbf{\ell}}_v, \ell_u),
\end{equation}
where $contexts(v)$ are context labels in the label walk starting from the target node $v$.
Finally, the model is trained to update $\mathbf{Z}$ and $\mathbf{W_{out}}$ to minimize the loss $\mathcal{L}$.
Remember that the context label is null if its corresponding node is not in the training set. 
We ignore the null label in the model training. 
In addition, when the context node is the same as the target node in a label walk, we exclude it from the loss computation to prevent information leaks.
Algorithm~\ref{algo:Labelembeddings} presents the details for generating label context embeddings.

\subsection{Training Classifier using Label Context Embeddings}
To perform node classification, we finally train a node classifier (LCC) that predicts the class label of the target node using its label context embeddings. 
Since different types of label walks capture different aspects of class label connectivity, we train the MLP classifier using the concatenation of node features and all label context embeddings obtained from different types of label walks. 

\begin{figure*}[t] 
        \centering
        \includegraphics[width=0.5\textwidth]{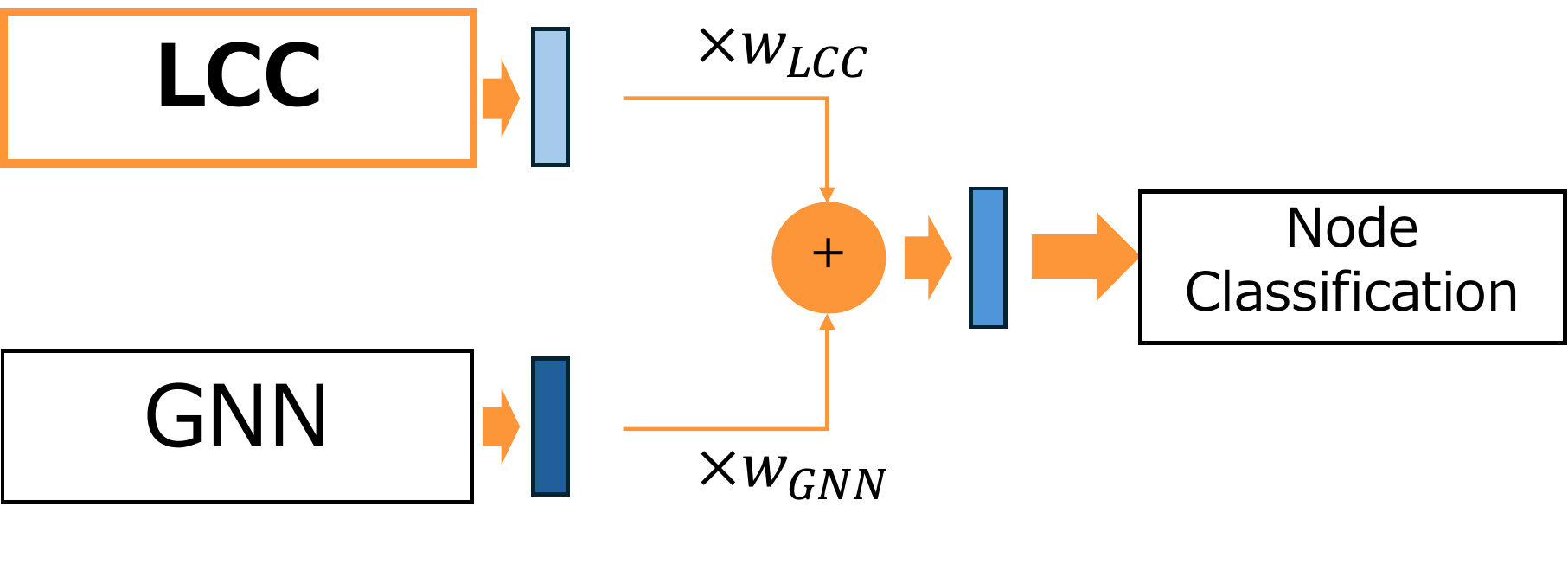}
        \label{fig:overview_integratte}
    \caption{
    Integration of LCC and any GNN. The final prediction is computed by average outputs from LCC and GNN weighted by their importance.
    }
    \label{fig:integration} 
\end{figure*}

\section{Integration of LCC and GNN}
\label{sec:integration}
LCC captures the higher-order class label connectivity that GNNs fail to capture.
Therefore, LCC complements the capability of GNNs. 
Since there are several GNNs designed for heterophilic graphs, such as H2GCN, LINKX, and GloGNN, we integrate LCC and one of these models and adaptively learn the importance of both LCC and GNN to achieve high accuracy.
An overview of integration is illustrated in Figure~\ref{fig:integration}. Since our integration does not need additional training for both LCC and GNN, it does not require additional training costs.

Specifically, after independently training both LCC and GNN on the training set, we determine the importance weights of LCC and GNN using validation loss.
The reason we use the validation loss is to prevent overfitting to the training data.
The losses $\mathcal{L}_{LCC}$ and $\mathcal{L}_{GNN}$ of each model are calculated as the cross-entropy loss $\mathcal{L}_{CE}$ between the predictions $\hat{y}_v^{LCC}$, $\hat{y}_v^{GNN}$ and the ground-truth label $y_v$ in the validation set.
\begin{equation}
    \mathcal{L}_{LCC}=\sum_{v \in V_{val}}\mathcal{L}_{CE}(\hat{y}_v^{LCC}, y_v), \label{eq:llcc}
\end{equation}
\begin{equation}
    \mathcal{L}_{GNN}=\sum_{v \in V_{val}}\mathcal{L}_{CE}(\hat{y}_v^{GNN}, y_v), \label{eq:lgnn}
\end{equation}

\begin{algorithm}[t]
    \caption{Integration of LCC and any GNN}
    \label{algo:EnsembleNodeClassification}
    \begin{algorithmic}[1] 
    \REQUIRE Adjacency matrix $\mathbf{A}$, Feature matrix $\mathbf{X}$, Class label $\mathbf{y}$, \\
    Trained node classification models ($\text{LCC}, \text{GNN}$), Temperature $T$
    \ENSURE Ensemble output $\mathbf{Y}^{GNN+LCC}$
    
    
    \STATE {\bf \#\#\# Calculate model outputs \#\#\#}
    \begin{align*}
    \mathbf{Y}^{LCC} &= \text{LCC}(\mathbf{A}, \mathbf{X}, \mathbf{y}), 
    \mathbf{Y}^{GNN} = \text{GNN}(\mathbf{A}, \mathbf{X}, \mathbf{y}).
    \end{align*}
    
    \STATE {\bf \#\#\# Calculate model weights \#\#\#}
    \STATE Calculate the validation loss of each model using Equations \ref{eq:llcc} and \ref{eq:lgnn}.
    \STATE Calculate temperature-adjusted model weights using Equations \ref{eq:wlcc} and \ref{eq:wgnn}.
    
    \STATE Calculate ensemble output using Equation \ref{eq:loss}.
    
    \RETURN $\mathbf{Y}^{GNN+LCC}$
    \end{algorithmic}
\end{algorithm}

Since models with lower validation loss are considered more reliable for node classification, we compute the importance weights $w_{LCC}$ and $w_{GNN}$ as the reciprocal of the validation loss as follows:
\begin{align}
    w_{LCC} &= \frac{\exp\left( \frac{1}{\mathcal{L}_{LCC}} \cdot \frac{1}{T} \right)}{\exp\left( \frac{1}{\mathcal{L}_{LCC}} \cdot \frac{1}{T} \right) + \exp\left( \frac{1}{\mathcal{L}_{GNN}} \cdot \frac{1}{T} \right)}, \label{eq:wlcc}\\
    w_{GNN} &= 1 - w_{LCC}, \label{eq:wgnn}
\end{align}
where $T$ is a temperature parameter to adjust the importance weights.

Finally, the prediction $\mathbf{Y}^{GNN+LCC}$ is computed by weighting the predictions $\mathbf{Y}^{LCC}$ and $\mathbf{Y}^{GNN}$ of the two models with their respective importance weights.
\begin{equation}
    \mathbf{Y}^{GNN+LCC} = w_{LCC} \cdot \mathbf{Y}^{LCC} + w_{GNN} \cdot \mathbf{Y}^{GNN} \label{eq:loss}
\end{equation}
In this way, we can integrate LCC and any GNN without additional training and effectively complement their limitations.
Algorithm~\ref{algo:EnsembleNodeClassification} presents this integration algorithm.

\section{Evaluation Experiments}
\label{sec:experiments}
We evaluate our proposal, the integration of GNN and LCC (GNN+LCC),  to answer the following four questions:
\begin{description}
    \item \textbf{Q$1$}: Does GNN+LCC contribute to improving the accuracy of existing GNN designed for heterophilous graphs?
    \item \textbf{Q$2$}: Are label walks effective for node classification?
    \item \textbf{Q$3$}: Which label walk types are important for node classification?
    \item \textbf{Q$4$}: Does the length of the label walk affect performance?
\end{description}
Our code can be found at \url{https://github.com/TakahashiTakutooo/GNN_LCC}.

\subsection{Experimental Setup}
\noindent
{\bf Datasets.}
We use seven heterophilous directed graph datasets: Cornell, Texas, Wisconsin, Chameleon, Squirrel, Roman-Empire, and Amazon-Ratings~\cite{Geom-GCN, Criticallook}.
Cornell, Texas, and Wisconsin represent category-based connections in university web pages.
Chameleon and Squirrel are networks focused on specific topics from Wikipedia.
Roman-Empire represents word connections in Wikipedia articles about the Roman Empire, while Amazon-Ratings represents product connections frequently purchased together on Amazon.
Table~\ref{datasets} shows the statistics of the datasets.
\begin{table}[t]
\centering
\caption{The statistics of datasets}
\label{datasets}
\scalebox{1}{
\begin{tabular}{l|rrrrr}  
\toprule
dataset & \#nodes  & \#edges & \#attributes & \#class & edge homophily \\
\midrule
Cornell & 183 & 298 & 1,703 & 5 & 0.131 \\
Texas & 183 & 325 & 1,703 & 5 & 0.108 \\
Wisconsin & 251 & 515 & 1,703 & 5 & 0.196 \\
Chameleon & 2,277 & 36,101 & 2,325 & 5 & 0.235 \\
Squirrel & 5,201 & 217,073 & 2,089 & 5 & 0.224 \\
Roman-Empire & 22,662 & 44,363 & 300 & 18 & 0.044 \\
Amazon-Ratings & 24,492 & 113,276 & 300 & 5 & 0.382 \\
\bottomrule
\end{tabular}
}
\end{table}


\noindent
{\bf Data Splitting.}
We split each dataset into training, validation, and test sets using five different split methods provided by PyG\footnote{\url{https://github.com/pyg-team/pytorch\_geometric}}.
For each split method, we perform node classification and compute the average accuracy.
The data split ratios are as follows:
For Cornell, Texas, Wisconsin, Chameleon, and Squirrel, the train/validation/test split is $48$\%/$32$\%/$20$\%.
For Roman-Empire and Amazon-Ratings, the split is $50$\%/$25$\%/$25$\%.

\noindent
{\bf Hyperparameters.}
For our proposed method, we perform a grid search using the validation set to determine the hyperparameters: label walk length, number of label walks, label context embedding dimension, and temperature parameter.
The search ranges for each parameter are as follows: label walk length: \{1, 2, 3\}, number of label walks: \{3, 5, 7\}, \{8, 16, 32\}, temperature parameter: \{0.01, 0.02, \dots, 0.09, 0.1, 0.2, \dots, 1.0\}.

\subsection{Experimental Results}
\subsubsection{\textbf{Q$1$}: Does GNN+LCC contribute to improving the accuracy of existing GNN designed
for heterophilous graphs?}
To evaluate whether LCC enhances existing GNNs for heterophilous graphs, we integrate LCC with H2GCN, LINKX, and GloGNN (denoted as H2GCN+LCC, LINKX+LCC, and GloGNN+LCC) and compare their performance.
Additionally, we compare with standard GNN baselines, GCN and GAT.
\begin{table*}[t]
    \centering
    \caption{Node classification accuracy (\%) on the test data.
For GNN+LCC (H2GCN+LCC, LINKX+LCC, GloGNN+LCC), the accuracy differences ($\Delta$) from the respective GNN model and LCC are also indicated. The highest accuracy for each dataset is highlighted in bold.}
    \label{table_results}
    \scalebox{0.94}{
        \begin{tabular}{l||ccccccc}
            \toprule
            & Cornell & Texas & Wisconsin & Chameleon & Squirrel & Roman- & Amazon- \\
            &  &  &  &  &  & Empire & Ratings \\
            \midrule[\heavyrulewidth]
            GCN & 43.24 \textsubscript{±4.8} & 56.22 \textsubscript{±4.6} & 53.33 \textsubscript{±4.8} & 62.76 \textsubscript{±0.50} & 44.36 \textsubscript{±2.7} & 47.24 \textsubscript{±0.84} & 45.92 \textsubscript{±0.96} \\
            GAT & 46.49 \textsubscript{±10} & 53.51 \textsubscript{±11} & 48.24 \textsubscript{±4.4} & 65.22 \textsubscript{±2.3} & 42.56 \textsubscript{±5.9} & 55.34 \textsubscript{±1.1} & 44.75 \textsubscript{±0.97} \\
            H2GCN & 74.05 \textsubscript{±4.3} & 80.54 \textsubscript{±7.5} & 80.00 \textsubscript{±2.8} & 67.46 \textsubscript{±2.0} & 55.41 \textsubscript{±2.0} & 79.09 \textsubscript{±0.65} & 45.58 \textsubscript{±0.13} \\
            LINKX & 66.49 \textsubscript{±9.1} & 64.32 \textsubscript{±5.5} & 79.22 \textsubscript{±6.1} & 63.99 \textsubscript{±0.7} & 58.94 \textsubscript{±2.6} & 52.35 \textsubscript{±0.71} & 53.70 \textsubscript{±1.8} \\
            GloGNN & 75.14 \textsubscript{±3.1} & 77.84 \textsubscript{±4.6} & \textbf{85.1 \textsubscript{±2.3}} & 68.2 \textsubscript{±1.5} & 58.48 \textsubscript{±0.78} & 58.24 \textsubscript{±0.87} & 49.94 \textsubscript{±0.44} \\
            \midrule
            LCC & 73.51 \textsubscript{±5.7} & 77.84 \textsubscript{±3.1} & 78.43 \textsubscript{±5.4} & 56.01 \textsubscript{±1.6} & 44.36 \textsubscript{±1.4} & 81.05 \textsubscript{±0.44} & 52.17 \textsubscript{±0.48} \\
            \midrule
            \textbf{H2GCN+LCC} & \textbf{78.92 \textsubscript{±2.0}} & \textbf{84.32 \textsubscript{±3.1}} & 83.53 \textsubscript{±4.7} & 68.25 \textsubscript{±1.2} & 55.14 \textsubscript{±1.0} & \textbf{84.10 \textsubscript{±0.41}} & 52.09 \textsubscript{±0.60} \\
            $\Delta$H2GCN & \textbf{+4.87} & \textbf{+3.78} & \textbf{+3.53} & \textbf{+0.79} & -0.27 & \textbf{+5.01} & \textbf{+6.51} \\
            $\Delta$LCC & \textbf{+5.41} & \textbf{+6.48} & \textbf{+5.10} & \textbf{+12.24} & \textbf{+10.78} & \textbf{+3.05} & -0.08 \\
            \midrule
            \textbf{LINKX+LCC} & 76.76 \textsubscript{±6.0} & 77.84 \textsubscript{±3.9} & 78.82 \textsubscript{±3.3} & 66.93 \textsubscript{±2.0} & \textbf{61.56 \textsubscript{±1.0}} & 81.11 \textsubscript{±0.15} & \textbf{57.03 \textsubscript{±0.28}} \\
            $\Delta$LINKX & \textbf{+10.27} & \textbf{+13.52} & -0.40 & \textbf{+2.94} & \textbf{+2.62} & \textbf{+28.76} & \textbf{+3.33} \\
            $\Delta$LCC & \textbf{+3.25} & 0.00 & \textbf{+0.39} & \textbf{+10.92} & \textbf{+17.20} & \textbf{+0.06} & \textbf{+4.86} \\
            \midrule
            \textbf{GloGNN+LCC} & 76.76 \textsubscript{±3.2} & 81.62 \textsubscript{±4.6} & \textbf{85.1 \textsubscript{±2.9} }& \textbf{69.96 \textsubscript{±2.1}} & 61.31 \textsubscript{±1.4} & 80.56 \textsubscript{±0.3} & 54.42 \textsubscript{±0.24} \\
            $\Delta$GloGNN & \textbf{+1.62} & \textbf{+3.78} & 0.00 & \textbf{+1.76} & \textbf{+2.83} & \textbf{+22.32} & \textbf{+4.48} \\
            $\Delta$LCC & \textbf{+3.25} & \textbf{+3.78} & \textbf{+6.67} & \textbf{+13.95} & \textbf{+16.95} & -0.49 & \textbf{+2.25} \\
            \bottomrule
        \end{tabular}
    }
\end{table*}

\noindent
{\bf Overall.}
Table~\ref{table_results} shows the experimental results.
The most important observation is that the highest node classification accuracy is achieved by one of GNN+LCC for all datasets.
In addition, integrating LCC and GNN actually improves accuracy compared to each model alone in most cases (see $\Delta$H2GCN, $\Delta$LINKX, $\Delta$GloGNN, $\Delta$LCC).
This result confirms that 1) LCC and GNN capture different graph properties and complement each other, and 2) our integration scheme effectively learns the importance weights of both LCC and GNN.

\noindent
{\bf Analysis for exceptional cases.}
We also observe that there are only four exceptional cases out of 21 cases.  
GNN alone achieves slightly higher accuracy than GNN+LCC in two cases, {\bf{case1}}: H2GCN+LCC for the Squirrel dataset (H2GCN is 0.27 higher) and {\bf{case2}}: LINKX+LCC for the Wisconsin dataset (LINKX is 0.40 higher). 
Also, LCC alone achieves slightly higher accuracy than GNN+LCC in two cases, {\bf{case3}}: H2GCN+LCC for the Amazon-Ratings dataset (LCC is 0.08 higher) and {\bf{case4}}: GloGNN+ LCC for the Roman-Empire dataset (LCC is 0.49 higher). 
The three cases above can be explained by investigating the importance weights between GNN and LCC: the integration does not work well when one of the model weights is extremely high.
Table~\ref{H2GCN+LCC_weight} shows the learned importance weights of LCC and H2GCN. The H2GCN weight (0.916) for the Squirrel dataset ({\bf{case1}}) and the LCC weight (0.925) for the Amazon-Ratings dataset ({\bf{case3}}) are extremely high. 
We observe the same phenomena for the LCC weight (0.881) for the Roman-Empire dataset ({\bf{case4}}). 
This result implies that there is room for further revision of the weight control mechanism.

\begin{table}[t]
\centering
\caption{The importance weights of each model and temperature parameter in H2GCN+LCC, optimized using grid search.}
\label{H2GCN+LCC_weight}
\resizebox{0.8\linewidth}{!}{%
\begin{tabular}{l||cc|c}
\toprule
& LCC weight & H2GCN weight & temperature parameter\\
\midrule
Cornell & 0.587 & 0.413 & 0.20 \\
Texas & 0.482 & 0.518 & 0.70 \\
Wisconsin & 0.632 & 0.368 & 0.30 \\
Chameleon & 0.102 & 0.898 & 0.09 \\
Squirrel & 0.084 & 0.916 & 0.08 \\
Roman-Empire & 0.502 & 0.498 & 1.00 \\
Amazon-Rationgs & 0.925 & 0.075  & 0.02 \\
\bottomrule
\end{tabular}%
}
\end{table}


\noindent
{\bf Importance weights of the two models in integration.}
Table~\ref{H2GCN+LCC_weight} shows the importance weights and temperature parameters for H2GCN+LCC. We omit the results for LINKX+LCC and GloGNN+LCC due to space limitations.
Overall, the importance weights are relatively balanced across datasets, except for the Chameleon, Squirrel, and Amazon-Ratings datasets.
These results confirm the effectiveness of our weight control mechanism for the LCC and GNN integration.
As examples of imbalanced weights, for the Chameleon and Squirrel datasets, H2GCN significantly outperforms LCC (see Table~\ref{table_results}), resulting in very high H2GCN weights (0.898, 0.916, respectively).
Similarly, for the Amazon-Ratings dataset, LCC significantly outperforms H2GCN, leading to a very high LCC weight (0.925).

\subsubsection{\textbf{Q$2$}: Are label walks effective for node classification?}

\begin{table*}[t]
    \centering
    \caption{Node classification accuracy (\%) on the test data for vanilla MLP and LCC. The higher accuracy for each dataset is highlighted in bold, and the accuracy difference between LCC and vanilla MLP is indicated as $gain$.}
    \label{table_mlp}
\scalebox{1.0}[1.0]{
\begin{tabular}{l|ccccccc}
\toprule
 & Cornell & Texas & Wisconsin & Chameleon & Squirrel & Roman- & Amazon-  \\ 
 &         &       &           &           &          & Empire & Ratings  \\ 
\midrule[\heavyrulewidth]
MLP & $72.97_{\pm4.5}$ & $76.22_{\pm5.7}$ & $77.25_{\pm3.6}$ & $44.04_{\pm2.7}$ & $31.24_{\pm1.3}$ & $64.98_{\pm0.23}$ & $41.00_{\pm0.41}$ \\
\midrule
LCC & $\mathbf{73.51_{\pm5.8}}$ & $\mathbf{77.84_{\pm3.2}}$ & $\mathbf{78.43_{\pm5.4}}$ & $\mathbf{56.01_{\pm1.6}}$ & $\mathbf{44.36_{\pm1.5}}$ & $\mathbf{81.05_{\pm0.45}}$ & $\mathbf{52.17_{\pm0.48}}$ \\
$gain$ & $\mathbf{+0.54}$ & $\mathbf{+1.62}$ & $\mathbf{+1.18}$ & $\mathbf{+11.97}$ & $\mathbf{+13.12}$ & $\mathbf{+16.07}$ & $\mathbf{+11.17}$ \\
\bottomrule
\end{tabular}
} 
\end{table*}

To verify the effectiveness of using label context embeddings, we compare the accuracy between a vanilla MLP and LCC. The MLP takes only node features as input whereas LCC additionally takes label context embeddings obtained from four different label walks.

Table~\ref{table_mlp} shows that LCC outperforms the vanilla MLP in all datasets. 
This result confirms that the label context embeddings successfully capture class label connectivity and the label walks contribute to improving the node classification accuracy.
In particular, the accuracy is improved by more than 10\% for the Chameleon, Squirrel, Roman-Empire, and Amazon-Ratings datasets.
In particular, the accuracy improvement is significant for the Roman-Empire dataset, more than 16\%, which may be due to the large number of classes and the low homophily ratio of the dataset.


\subsubsection{\textbf{Q$3$}: Which label walk types are important for node classification?}
To evaluate the importance of each of the four types of label walks, we compare LCC variations which concatenate node features with the label context embedding generated from a single type of label walk: LCC\_Forward (using forward walks), LCC\_Backward (using backward walks), LCC\_Sibling (using sibling walks), and LCC\_Guardian (using guardian walks).
We also compare with the vanilla MLP as a baseline. 

\begin{figure}[t]
    \centering
    \includegraphics[keepaspectratio, scale=0.30]{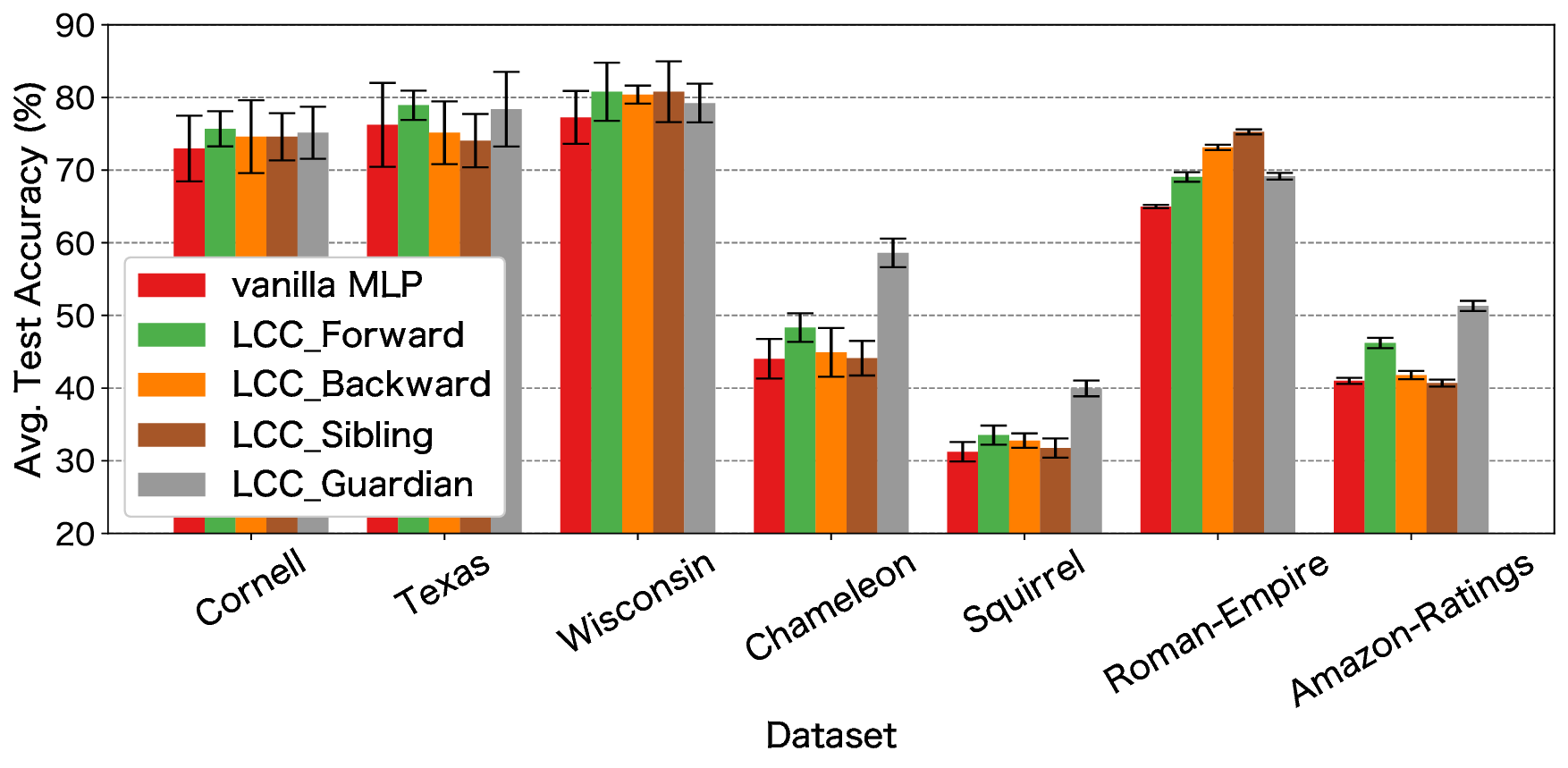}
    \caption{Node classification accuracy of LCC variations using each label walk type.}
    \label{fig:walk_type}
\end{figure}

Figure~\ref{fig:walk_type} shows that the most important label walk varies across datasets.
It is quite interesting to observe that, for the Chameleon, Squirrel, and Amazon-Ratings datasets, the guardian walks contribute the most to improving accuracy and the gain is quite significant.
Since the guardian walk captures class label connectivity between guardian nodes, we conjecture that these datasets exhibit strong such connectivity.
In contrast, for the Roman-Empire dataset, the sibling walks contribute the most to improving accuracy.
The result confirms the significant effectiveness of using different label walks, particularly the guardian walk and sibling walk.

\subsubsection{\textbf{Q$4$}: Does the length of the label walk affect performance?}  
To investigate the impact of label walk lengths, 
we examine the performance of LCC\_Forward, LCC\_Backward, LCC\_Sibling, and LCC\_Guardian by varying the label walk length to $1$, $2$, and $3$.

\begin{figure}[t]
\centering
\includegraphics[keepaspectratio, scale=0.5]{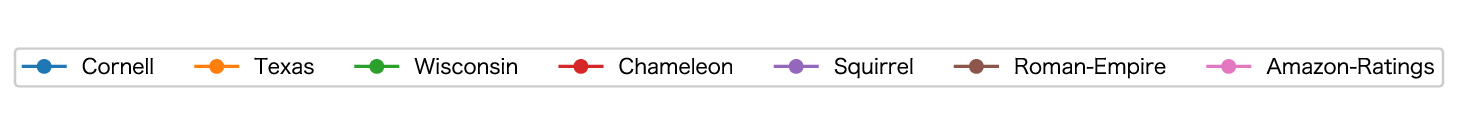}\\[1ex]

\begin{tabular}{cc} 
    \begin{minipage}{0.5\hsize}
    \centering
    \includegraphics[keepaspectratio, scale=0.3]{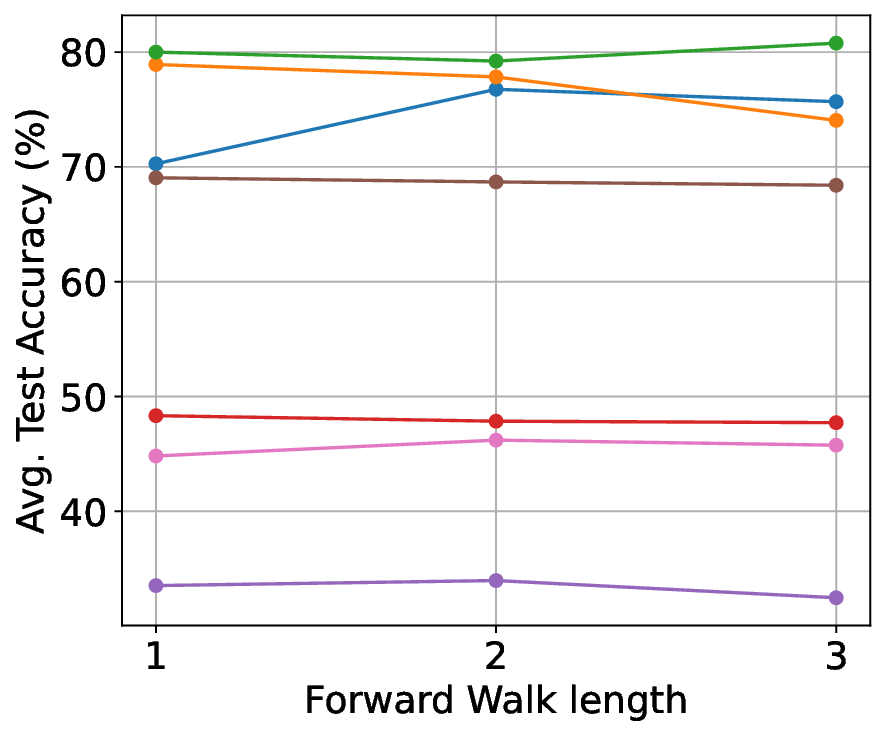}\\
    \text{(a) LCC\_Forward}
    \end{minipage} &
  
    \begin{minipage}{0.5\hsize}
    \centering
    \includegraphics[keepaspectratio, scale=0.3]{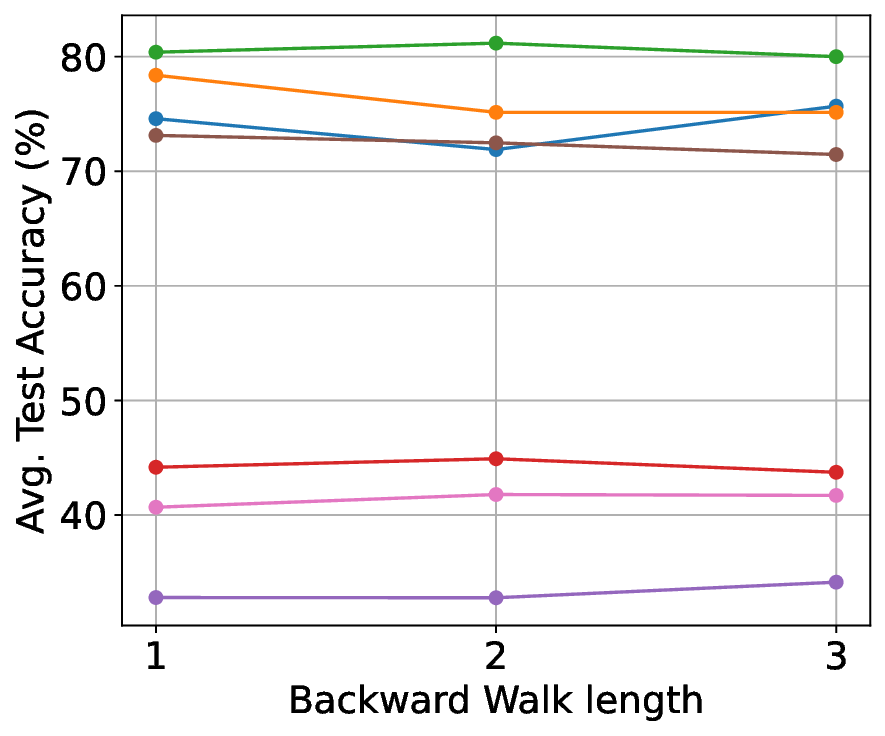}\\
    \text{(b) LCC\_Backward}
    \end{minipage} \\

    \begin{minipage}{0.5\hsize}
    \centering
    \includegraphics[keepaspectratio, scale=0.3]{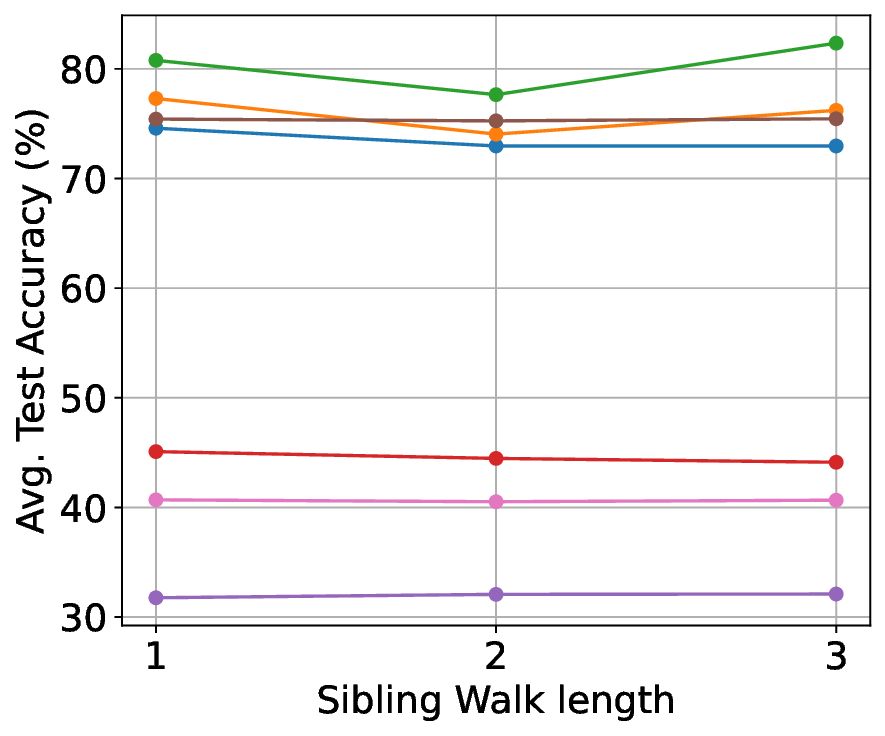}\\
    \text{(c) LCC\_Sibling}
    \end{minipage} &

    \begin{minipage}{0.5\hsize}
    \centering
    \includegraphics[keepaspectratio, scale=0.3]{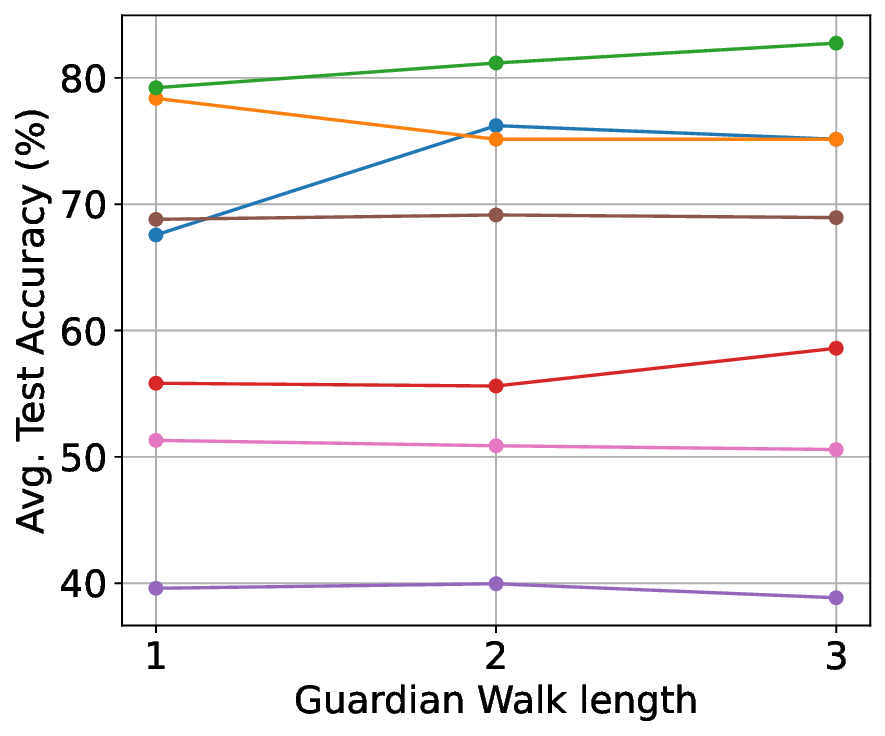}\\
    \text{(d) LCC\_Guardian}
    \end{minipage}  
\end{tabular}
 
\caption{Node classification accuracy (y-axis) when we change label walk length (x-axis).}
\label{fig:labelwalk_length}
\end{figure}

Overall, the result in Figure~\ref{fig:labelwalk_length} indicates that the higher-order class label connectivity is crucial for improving the accuracy, as demonstrated by performance improvements when we increase the walk length for forward/backward walks (the class label connectivity order is walk length) or we use sibling/guardian walks (the class label connectivity order is 2).
Also, the trends vary across datasets and the type of label walks, highlighting the importance of tuning the label walk length as a hyperparameter using a validation set.

For forward walks in the Cornell dataset, the accuracy is significantly improved when increasing the walk length from $1$ to $2$ and $3$, while in the Texas dataset, the accuracy is decreased significantly when the walk length is increased to $3$.  
For backward walks in the Squirrel and Amazon-Ratings datasets, the accuracy is improved with longer walk lengths, whereas the accuracy is decreased for the Texas and Roman-Empire datasets.
For sibling walks, the accuracy of Wisconsin drops significantly when increasing the walk length from $1$ to $2$ but improves again when the walk length is increased to $3$.  
For guardian walks, the accuracy improves with longer walk lengths in the Cornell, Wisconsin, and Chameleon datasets.

\section{Conclusion}
\label{sec:conclusion}
In this paper, we focused on the fact that conventional GNNs are designed in a way that does not effectively utilize structural information between class labels.
The proposed method consists of two key components: 
(1) the development of a node classifier, LCC, which estimates the class label of a target node based on the neighboring class labels that constitute directed label walks, and 
(2) the integration of LCC with GNN, enabling both models to complement each other by capturing different graph characteristics.
The experiments demonstrated that the proposed method improves node classification accuracy across seven heterophilous graphs.

\begin{credits}
\subsubsection{\ackname} This work was supported by JSPS KAKENHI Grant Numbers JP20H00583 and JP25H01117 and JST ASPIRE Grant Number JPMJAP2328.
\end{credits}
%
%
%
\footnotesize

%
\appendix

\section{Analysis of Sibling Walk and Guardian Walk}
\begin{figure}[h!]
    \centering
    \begin{minipage}{0.33\textwidth}
        \centering
        \includegraphics[width=\textwidth]{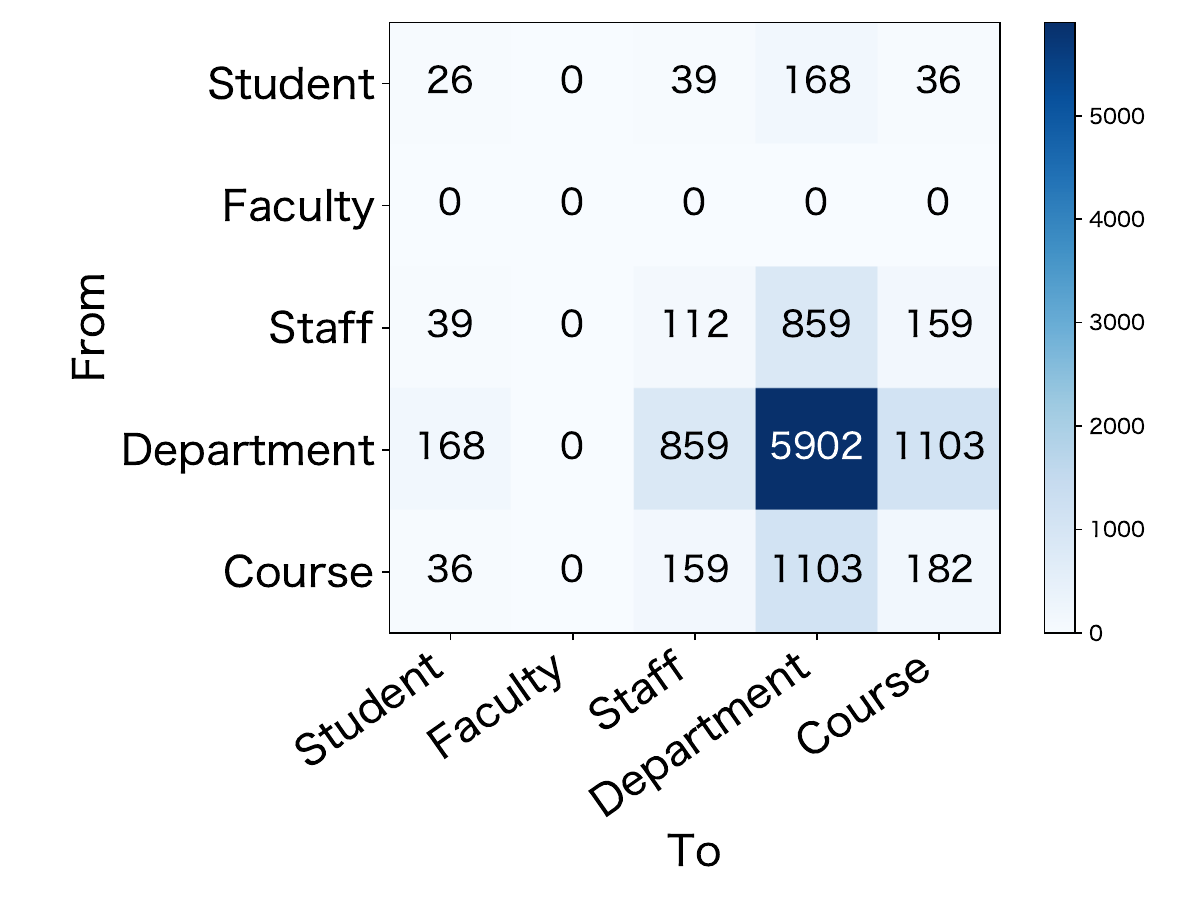}
        \subcaption{Sibling class label connectivity}
    \end{minipage}
    \begin{minipage}{0.33\textwidth}
        \centering
        \includegraphics[width=\textwidth]{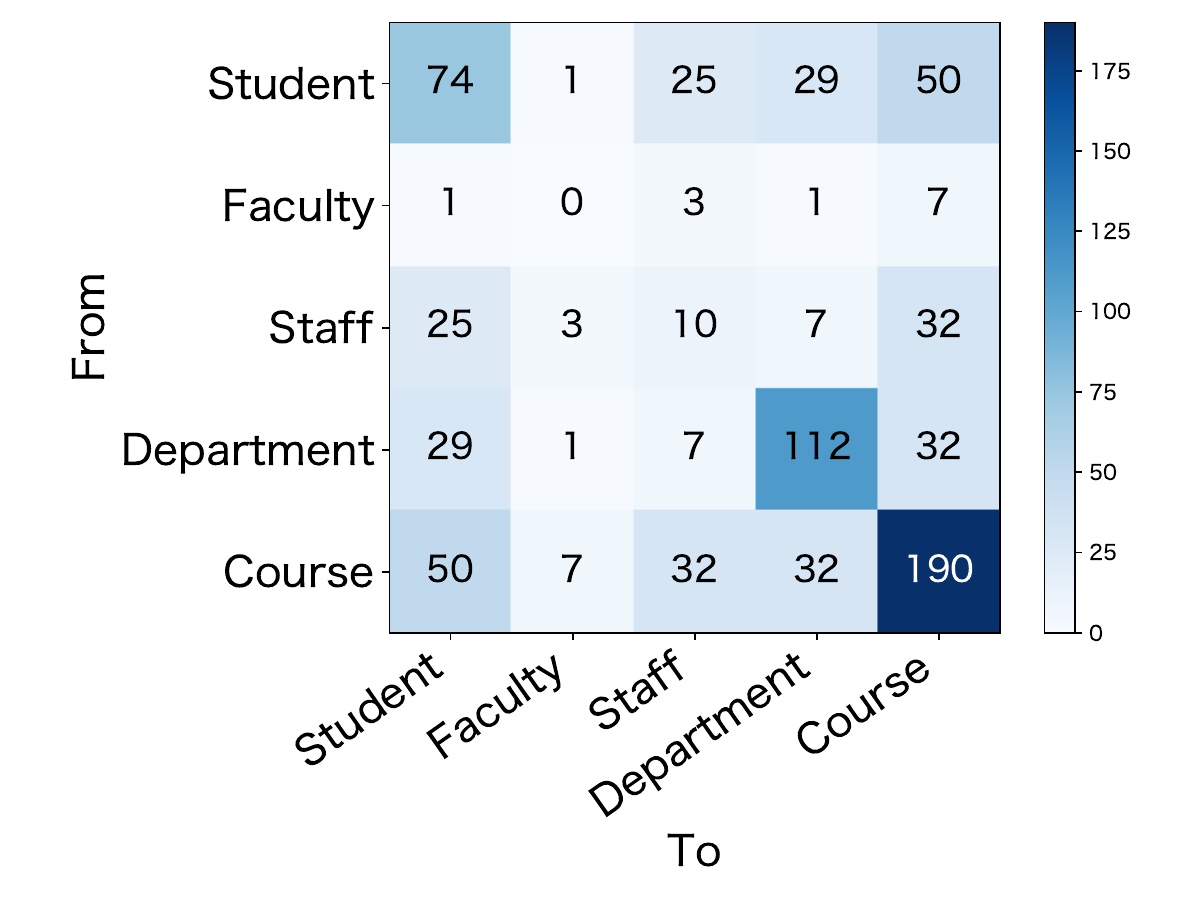}
        \subcaption{Guardian class label connectivity}
    \end{minipage}
    \caption{The class label connectivity in the Texas dataset. (a) sibling class label connectivity from class label in y-axis to class label in x-axis. (b) guardian class label connectivity from class label in y-axis to class label in x-axis. }
    \label{fig:texas_sibling_guardian}
\end{figure}

Figure~\ref{fig:texas_sibling_guardian} illustrates the sibling and guardian class label connectivity in the heterophilous directed graph of the Texas dataset.
Figure~\ref{fig:texas_sibling_guardian} (a) shows the sibling class label connectivity. 
According to this, Department nodes have strong connectivity to Department nodes, while nodes of other class labels also show relatively strong connectivity to Department nodes.
In contrast, Figure~\ref{fig:texas_sibling_guardian} (b) shows the guardian class label connectivity, revealing that nodes with Student nodes, Department nodes, and Course nodes have strong connectivity to the same class label nodes.
This indicates that guardian walk is crucial for the Texas dataset.
Furthermore, our experimental results (Q3) corroborate the effectiveness of guardian walk on this dataset.

We also report the 1st-order connectivity and the sibling and guardian class label connectivity for other datasets, Cornell, Wisconsin, Chameleon, Squirrel, Roman-Empire, and Amazon-Ratings in the Supplementary Material.


\section{The Best Hyper Parameters of GNN+LCC}

For our proposed method, we perform a grid search using the validation set to determine the hyperparameters: label walk length, number of label walks, label context embedding dimension, and temperature parameter.
The search ranges for each parameter are as follows: label walk length: \{1, 2, 3\}, number of label walks: \{3, 5, 7\}, \{8, 16, 32\}, temperature parameter: \{0.01, 0.02, \dots, 0.09, 0.1, 0.2, \dots, 1.0\}.
The best hyperparameters for each dataset are reported in Table~\ref{tab:best_hyparameters}.

\begin{table}[h]
    \centering
    \caption{The best hyperarameters.}
    \label{tab:best_hyparameters}
    \begin{subtable}{\textwidth} 
        \centering
        \caption{Label Walk parameters (Forward, Backward, Sibling, Guardian) of LCC}
        \label{tab:sub_lcc_walk_parameters_combined}
        \begin{tabular}{@{}l |rrr rrr rr rr@{}}
        \toprule
        & \multicolumn{3}{c}{Forward} & \multicolumn{3}{c}{Backward} & \multicolumn{2}{c}{Sibling} & \multicolumn{2}{c}{Guardian} \\
        \cmidrule(lr){2-4} \cmidrule(lr){5-7} \cmidrule(lr){8-9} \cmidrule(lr){10-11}
                  & Len. & Num. & Dim. & Len. & Num. & Dim. & Len. & Dim. & Len. & Dim. \\ 
        \midrule
        Cornell   & 3 & 5 & 8 & 1 & 3 & 8 & 1 & 8 & 3 & 8 \\
        Texas     & 1 & 7 & 8 & 2 & 7 & 8 & 2 & 16 & 1 & 8 \\
        Wisconsin & 3 & 5 & 32 & 1 & 5 & 16 & 1 & 16 & 1 & 16 \\
        Chameleon & 1 & 5 & 8 & 2 & 3 & 8 & 3 & 8 & 3 & 16 \\
        Squirrel  & 1 & 3 & 8 & 2 & 7 & 16 & 1 & 32 & 2 & 32 \\
        Roman-Empire & 1 & 3 & 32 & 1 & 3 & 16 & 2 & 16 & 2 & 16 \\
        Amazon-Ratings & 2 & 5 & 8  & 2 & 7 & 8 & 1 & 16 & 1 & 8 \\
        \bottomrule
        \end{tabular}
    \end{subtable}
    \vspace{1.5em}
    \begin{subtable}{0.7\textwidth}
        \centering
        \caption{Best temperature parameters for GNN + LCC models}
        \label{tab:sub_temperature_parameters_corrected}
        \begin{tabular}{@{}l|rrr@{}}
        \toprule
        & H2GCN + LCC & LINKX + LCC & GloGNN + LCC \\
        \midrule
        Cornell & 0.2 & 1.0 & 1.0 \\
        Texas & 0.7 & 0.3 & 1.0 \\
        Wisconsin & 0.3 & 1.0 & 1.0 \\
        Chameleon & 0.09 & 0.05 & 0.1 \\
        Squirrel & 0.08 & 0.1 & 0.2 \\
        Roman-Empire & 1.0 & 0.09 & 0.2 \\
        Amazon-Ratings & 0.02 & 0.2 & 0.1 \\
        \bottomrule
        \end{tabular}
    \end{subtable}
\end{table}

\end{document}